\documentclass[letterpaper, 10 pt, conference]{ieeeconf}  

\usepackage{amsmath,amsfonts,amsthm}
\usepackage{algorithm}
\usepackage[noend]{algpseudocode}
\usepackage{graphicx}
\usepackage{multirow}
\usepackage{makecell}
\usepackage{caption}
\usepackage{subcaption}

\IEEEoverridecommandlockouts                              

\title{\LARGE \bf
A Safe Hierarchical Planning Framework for Complex Driving Scenarios based on Reinforcement Learning
}

\author{Jinning Li, Liting Sun, Jianyu Chen, Masayoshi Tomizuka and Wei Zhan
\thanks{J. Li, L. Sun, M. Tomizuka and W. Zhan are with the Department of Mechanical Engineering, University of California, Berkeley, CA 94720, USA
        {\tt\small 	\{jinning\_li, litingsun, tomizuka, wzhan\}@berkeley.edu}}%
\thanks{J. Chen is with the Institute for Interdisciplinary Information Sciences, Tsinghua University, Beijing, 100084, China
        {\tt\small	jianyuchen@tsinghua.edu.cn}}%
}

\begin{document}

\maketitle
\thispagestyle{empty}
\pagestyle{empty}

\begin{abstract}

Autonomous vehicles need to handle various traffic conditions and make safe and efficient decisions and maneuvers. However, on the one hand, a single optimization/sampling-based motion planner cannot efficiently generate safe trajectories in real time, particularly when there are many interactive vehicles near by. On the other hand, end-to-end learning methods cannot assure the safety of the outcomes. To address this challenge, we propose a hierarchical behavior planning framework with a set of low-level safe controllers and a high-level reinforcement learning algorithm (H-CtRL) as a coordinator for the low-level controllers. Safety is guaranteed by the low-level optimization/sampling-based controllers, while the high-level reinforcement learning algorithm makes H-CtRL an adaptive and efficient behavior planner. To train and test our proposed algorithm, we built a simulator that can reproduce traffic scenes using real-world datasets. The proposed H-CtRL is proved to be effective in various realistic simulation scenarios, with satisfying performance in terms of both safety and efficiency.

\end{abstract}

\section{INTRODUCTION}
\begin{figure}
     \centering
     \includegraphics[width=\linewidth]{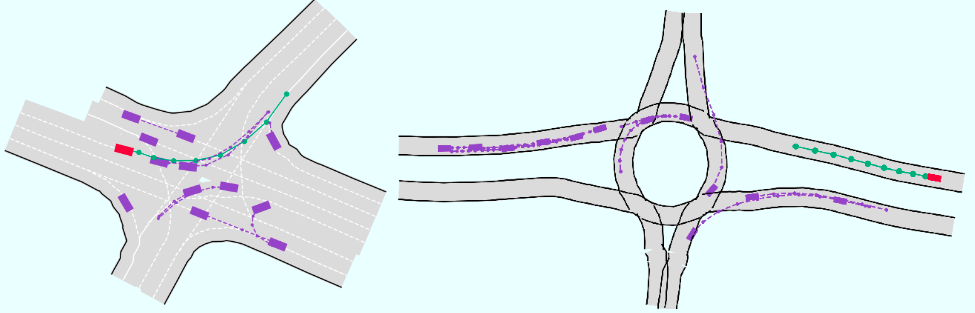}
     \caption{Various complex real-world traffic conditions in which autonomous vehicles (red) must drive safely and efficiently.}
     \label{fig:wholemap}
     \vspace{-6mm}
\end{figure}

Recently the field of autonomous driving has witnessed rapid development. A number of novel behavior planning algorithms have been proposed, many of which are awesome achievements. However, it is almost impossible for a single behavior planner to drive through so many different real and complex scenarios. For example, one needs to ride in an autonomous car in his everyday life. The autonomous car should be able to drive both in cities and on highways, which include various intersections, roundabouts, and merging and following scenarios (as shown in Figure \ref{fig:wholemap}). One single planning algorithm, such as an optimization-based planner, may never find its solution in each complex traffic condition. The hyperparameters in each planner or policy may work well in intersection scenarios, but it is very likely to fail in highway merging scenarios, because the optimal driving settings in each scenario, e.g. the optimal speed and spacing, are so difficult to adjust with only one fixed planner. 

The most difficult part in this complex driving problem lies in the rapidly changing environment. The road curvature may change, and the number of obstacles may vary in each of these self-driving scenarios. For example, when an autonomous vehicle is trying to make a left turn at an intersection, it has to consider the number of oncoming vehicles, the actual speed of them, and their distance to the intersection. However, when the self-driving vehicle is finding its way to merge to the next lane on the highway, although it also has to analyze similar information, it needs to do it in a whole different larger scale for distance, velocity, etc. We therefore have to find an adaptive policy that can make high-quality decisions in various scenarios. 

Each behavior and motion planner from existing work has its own advantages and disadvantages. For example, learning-based methods can recognize the specific properties of different scenarios with less effort, but it is hard to interpret the results and generalize them to other scenarios. On the other hand, non-learning-based methods can ensure the safety of the agent and deal with many similar cases without too much modification, but it is difficult to tune the hyperparameters (e.g., the detailed cost function of motion planning) to be compatible for various traffic scenarios with different number of interactive agents. Therefore, based on the analysis of these limitations, we seek to design a hierarchical model to exploit the advantages of both reinforcement learning and non-learning-based policies. 

In this paper, we make the following contributions:
\begin{itemize}
    \item We propose a \textbf{H}ierarchical behavior planning framework with low-level safe \textbf{C}on\textbf{t}rollers and a high-level \textbf{R}einforcement \textbf{L}earning \textbf{(H-CtRL)}. It is an adaptive behavior planner that can make high-quality decisions in many complex traffic scenarios.
    \item A simulator that could reproduce traffic scenes from real-world traffic datasets is constructed, so that the proposed method can be trained and tested in realistic scenarios. 
    \item We test the proposed method H-CtRL in real-world traffic conditions, and it was proved to be capable of handling different planning tasks in various scenarios. 
\end{itemize}

\section{RELATED WORK}
\subsection{Non-Learning Based Methods}
Non-learning-based decision-making modules are considered to be safe-guaranteed and interpretable \cite{mcnaughton2011motion, paden2016survey}. But these planners tends to be too conservative. 
In \cite{defensive, bayesianPersuasive}, this problem was addressed by making the autonomous agent more cognizant of and reactive to obstacles. 
The authors of \cite{leverageHuman} proposed another framework leveraging effects on human actions to make it more interactive.
However, the methods aforementioned suffer more or less from their long computation time. Constrained Iterative Linear Quadratic Regulator \cite{CILQR} significantly reduced the operation time while preserving the safety. We try to inherit the safe guarantees and fast operation time from these methods, and select non-learning-based planners as our low-level safe controllers.

\subsection{Supervised Learning}
The application of supervised learning on autonomous driving dates back to the work in \cite{Alvinn}. The authors of \cite{end2endforAV} then learned a map from raw pixels directly to steering commands, where the concept of imitation learning (IL) began to surface. A more robust perception-action model was developed in \cite{endtoendPerception}. To enhance the safety of IL, \cite{sun2018fast} proposed a hierarchical framework which utilized a high-level IL policy and a low-level MPC controller to improve efficiency and safety. Similarly, to make IL generalizable and deal with complex urban scenarios, the authors of \cite{DILinurban} learned policies from offline connected driving data, and integrated a safety controller at test time. 

\subsection{Reinforcement Learning}
Reinforcement learning (RL) has also been extensively explored in autonomous driving. The algorithm in \cite{DRLframework} adopted Recurrent Neural Networks for information integration, and learned an effective driving policy on simulators. The work in \cite{virtualtoreal} developed a realistic translation network to make sim2real possible. \cite{scalable, interpretable} developed robust policies to make self-driving cars capable of driving through complex urban scenarios. One can also consider to incorporate prediction models \cite{evolvegraph, conditional} to build model-based RL planners.
We believe RL is a suitable high-level policy candidate. It can learn from experience to know which low-level controller is the most suitable at a specific time step.


Hierarchical Reinforcement Learning (HRL) can make the learning process more sample-efficient. 
The idea is to reuse the well trained network of one sub-goal on other similar tasks in HRL \cite{HRL, towardshrl}. 

There are also many variants of HRL. The work in \cite{learningHRL} integrated a sampling-based motion planner with a high-level RL policy, which can solve long horizon problems. Similarly, \cite{hoel2019combining} combined deep reinforcement learning with Monte Carlo sampling to achieve tactical decision making for autonomous vehicles. Authors of \cite{thananjeyan2020safety} developed SAVED which augmented the safety of model-based reinforcement learning with Value estimation from demonstrations. Only to plan in normal scenarios is not enough for reliable self-driving cars, so the authors of \cite{rl-il} developed a hybrid method with RL and IL policies to plan safely in near accident scenarios. The authors of \cite{socialAttention} proposed an attention-based architecture that can deal with a varying number of obstacles and the interaction in between. 

We adopted the basic ideas to reuse low-level controllers, and aimed to design a novel planning module that works in various traffic conditions. The high-level RL was trained to recognize and react to different environments, while the low-level conventional controllers fulfill the goals sent from the high-level RL and guarantee the safety in the same time.

\section{Problem Statement}
Throughout the paper, we focus on the behavior planning problem in different complex urban traffic scenarios. There is one ego agent and many other obstacle cars in the environment. Each of them has its own behavior pattern. Thus, we need a mechanism to model the evolution of each scene and the interactions among agents. We formulate the problem as a Partially Observable Markov Decision Process (POMDP). A POMDP can be defined as a tuple: $<\mathcal{S}, \mathcal{A}, \mathcal{O}, \mathcal{T}, \mathcal{Z}, \mathcal{R}, \gamma>$. $\mathcal{S}$ denotes the state space and $s\in \mathcal{S}$ is a state of the environment. $\mathcal{A}$ is defined to be the action space and $a \in \mathcal{A}$ is an action taken by the ego agent. $o \in \mathcal{O}$ is an observation received by the ego agent. The transition model $\mathcal{T}(s, a, s')$ is the probability of the transition from a current state - action pair $(s, a)$ to $s'$ at the next time step. $\mathcal{Z}(s, o)$ denotes the transition model, which calculates the probability of ending in the observation $o$ given a state $s$. The reward function is defined by $\mathcal{R}(s, a)$, which yields a specific reward via a state - action pair $(s, a)$. The discount factor is denoted by $\gamma$. The overall objective is to maximize the expected discount reward and find the corresponding optimal policy 
\begin{equation}
    \pi^* =\arg \max_{\pi} \mathbb{E}\left[ \sum_{t = 0}^{\infty}\gamma^t \mathcal{R}(s^t, \pi(o^t)) \right]
\end{equation}
where $s^t, \; o^t $ are the state and the observation at timestep $t$ of the environment, respectively. 

\section{METHOD}
\subsection{H-CtRL Framework}
\begin{figure}[h]
    \vspace{3pt}
     \centering
     \includegraphics[width=0.27\textwidth]{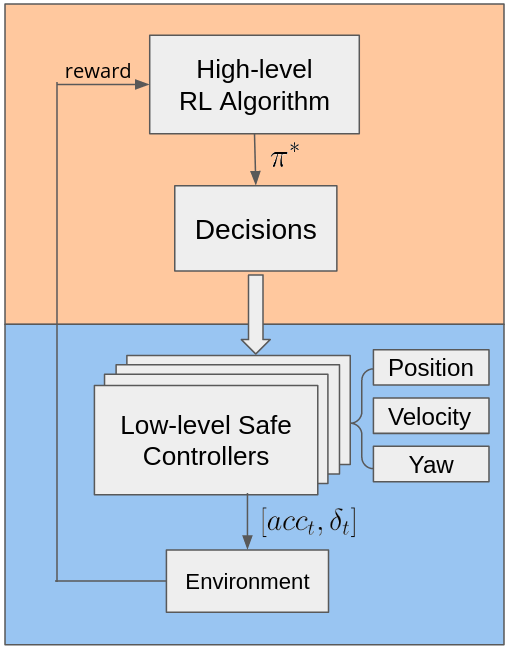}
     \caption{The hierarchical framework with low-level safe controllers and a high-level RL algorithm.}
     \label{fig:pipeline}
\end{figure}

We propose to solve the POMDP aforementioned with a hierarchical behavior planning framework H-CtRL (shown in Figure \ref{fig:pipeline}). It can be viewed as an integrated solver for the POMDP. 
We input the current state of the environment into the framework and then it outputs actions to be executed by the ego agent. 
The hierarchical framework consists of a collection of low-level controllers with safety constraints and a high-level Reinforcement Learning policy to manage them all. We aim to find an optimal high-level policy $\pi^*$ that can take advantage of one controller at a specific time step when doing behavior planning for the autonomous agent. 

In details, the state $s_t$ of the problem at the time step $t$ is designed to be the low-dimensional states of all agents presented in the environment at $t$, namely, 
\begin{equation}
    \begin{aligned}
    &s_t = \left[ s_t^1 \;\; || \;\; s_t^2 \;\; || \;\; \dots \;\; || \;\; s_t^m \right]\\
    &s_t^i = \begin{bmatrix}x_t^i & y_t^i & v_t^i & \theta^i \end{bmatrix}, \; i \in [1, m]
    \end{aligned}
\end{equation}

where $s_t^i,\; i \in [0, m]$ is the low-dimensional state of the $i$-th agent in the environment at time step $t$ (note that $s_t^0$ denotes the state of the ego agent), and $s_t$ is the state of the environment. The operator $||$ here is the concatenation operator. The action at time step $t$ is denoted by $a_t = [acc_t \;\; \delta_t]$ where $acc_t$ and $\delta_t$ are the acceleration and the steering angle of the ego agent, respectively. We choose the states of the $k$ nearest neighbors around the ego vehicle at time step $t$ to be the observation, namely, $o_t = [o_t^1 \;\; || \;\; o_t^2 \;\; || \;\; \dots \;\; || \;\; o_t^k]$ where $o_t^i, \; i\in [1, k]$ is the state of the $i$-th nearest obstacle around the ego car. We refer \cite{bicyclemodel} and select the bicycle model as the transition model for the ego agent. As for the other vehicles in the scene, their states would evolve according to the definition by the environment or the simulator.

\textbf{Low-Level Safe Controllers} consist a set of $n$ non-learning-based controllers, denoted by $\{ f_1, f_2, \dots, f_n\}$. They are like the workers within the hierarchical framework, who are responsible for specific tasks assigned by high-level coordinators. Each $f_i, \; i \in [1,n]$ has its own behavior pattern, either cooperative or selfish, either defensive or aggressive. Although they can lead to different driving strategies, they all have their own safety constraints when performing motion planning. For example, sampling-based planners would assign little probability to the area where an accident is more likely to happen, and optimization-based controllers would have huge cost in the objective function when the output lands in the dangerous area. Since the chosen low-level controller is the one whose action directly affects the evolution of the environment, these safety constraints of the low-level controllers are generally a good property that guarantees the safety of the whole hierarchical behavior planning framework. Specifically, the low-level controller takes in the observation $o_t$  and gives back an action $a_t$ at the time step $t$.

\begin{algorithm}
\caption{H-CtRL Training Algorithm}\label{alg:hctrl}
\begin{algorithmic}[1]
\State \textbf{Initialize} the simulation environment $env$ and get an initial observation $o_t$.
\State \textbf{Initialize} the policy and the target Q-network with weights $\theta$ and $\theta^-$. Set $\theta^- \gets \theta$.
\State \textbf{Instantiate} an empty reply buffer $\mathcal{B}$ with a maximum length of $l_B$.
\For {$h \gets 0$ \textbf{to} $N$ steps}
    \State Select an action $a_h \gets \arg \max_{a_h} Q_\theta (o_h, a_h)$ according to the $\epsilon-greedy$.
    \State Given $o_h$, the chosen low-level controller corresponding to $a_h$ acts $p$ timesteps in $env$ to get the next observation $o_{h+1}$ and the reward $r_h$.
    \State Push the transition $\mathcal{T} = \begin{bmatrix} o_h & a_h & r_h & o_{h+1} \end{bmatrix}$ into $\mathcal{B}$.
    \State Update the weights $\theta$ of the policy Q-network using the replay buffer $\mathcal{B}$.
    \If {$h$ \textbf{mod} $target\_update\_frequency == 0$}
        \State $\theta^- \gets \theta.copy()$
    \EndIf
    \If {the episode is done}
        \State Record the cumulative reward in this episode.
        \State Reset $env$ and get an initial observation $o_{t+1}$.
    \EndIf
\EndFor
\end{algorithmic}
\end{algorithm}

\textbf{The High-Level Reinforcement Learning} policy is the coordinator in this framework. It makes observations from the environment and gets to choose one of the low-level controllers that is the most suitable given the current observation. Therefore, the action space of the high-level RL is reduced and discretized from the original continuous space to a finite set of low-level controller's id $\{ 1, 2, \dots, n \}$. The actions of RL can be viewed as intermediate actions within the hierarchical framework, whereas the final output actions that directly interact and affect the environment are the outputs of low-level controllers. Here, we should note that to reduce the cardinality of the actions within one episode and improve the stability of the algorithm, the high-level RL switches its choice of low-level controllers every $p$ time steps, which means within the $p$ time steps of the environment, only one chosen low-level controller would plan trajectories consistently and would not be disabled by the RL. We therefore introduce the new timestep $h$ for the high-level RL: the original time step $t=h \cdot p$ should coincide with the time step $h$ in RL. 

The state and the observation at each time step of the RL problem are defined accordingly as $s_h$ and $o_h$ as aforementioned. One way to get $s_h$ and $o_h$ is to convert the time scale in RL back to the original scale, thus $s_h$ and $o_h$ should be the state and the observation at time step $t=h \cdot p$ in the environment. The transition model of RL is also defined within the new time scale, namely, $\mathcal{T}(s_h, a_h, s_{h+1})$, where $s_{h+1}$ is the next state of $s_h$ after applying $p$ actions by the low-level controller corresponding to $a_h$. Because of the existence of low-level controllers, we no longer need to worry about various tedious design details of the reward function $\mathcal{R}(s_h, a_h)$, and can simply adopt a very high-level one that encourages the completion of the planning task as fast as possible without any collision. In detail, the ego agent receives a positive reward that is proportion to its progress along its reference trajectory and a negative reward if the episode terminates early because of collision or low-level controller failure or other factors. 

Generally, the high-level RL can be trained using any model-free RL algorithms. In this paper, we choose to use Double Deep Q-Network (DDQN) in \cite{ddqn} to learn our high-level RL policy, for it is more stable and has less variance. The pseudo-code for the hierarchical planning framework is shown in algorithm \ref{alg:hctrl}.

\section{EXPERIMENTS}

\subsection{Simulator}
We constructed our own simulation environment based on the INTERACTION Dataset \cite{interactiondataset} and the OpenAI GYM toolkit. The road maps in the simulator were loaded from the INTERACTION Dataset map collections, which contained various real-world traffic scenarios recorded from many different countries. After the simulator finished constructing the road map, we specify an initial timestep from where vehicles data were loaded. The states of these vehicles other than the ego agent were all loaded from the dataset at each time step. Since these vehicle data were all collected from real-word traffics, we could simulate relatively realistic traffic conditions. The ego agent in the simulator would then be our self-driving car equipped with H-CtRL. The transition model of the ego agent was the bicycle model. When running experiments in the simulator, we input into it one low-level action, $a_t = [acc_t, \; \delta_t]$, and then it would take one step and output a reward, an observation, and a boolean indicating whether the episode had terminated, according to the bicycle model and the dataset. 

\subsection{Scenarios}
We consider two road maps from the INTERACTION dataset, and design different driving tasks in each of them. 
\begin{itemize}
    \item \textbf{TC\_BGR\_Intersection\_VA (VA)}. It is a map of a busy and complex intersection, which makes it difficult for the ego agent to avoid collisions.
    \item \textbf{DR\_USA\_Roundabout\_SR (SR)}. The map is collected from a real-world roundabout. The map is bigger than the previous intersection map and thus is difficult to navigate through. 
\end{itemize}
Since there are four directions in both scenarios, we design similar tasks in each of them. The ego agent should navigate through the traffic safely to make unprotected left turns, unprotected right turns, and straight crossings. By unprotected left or right turns, we mean that one must yield to other vehicles when turning left on green lights and turning right on red lights according to the traffic rules. 

\subsection{Low-level Controllers}
Generally speaking, we can choose any mature non-learning-based planner to make the proposed hierarchical framework inclusive and powerful. In this paper, we consider $n=9$ different Constrained Iterative Linear Quadratic Regulators (CILQR) \cite{CILQR} as the set of low-level controllers. 
The objective of CILQR is to find an optimal control sequence, namely, an optimal action sequence $a^*$ given an initial observation $o_0$ that minimizes a cost function:
\begin{equation}
    \begin{aligned}
    &a^*, o^* = \arg \min_{a, o} \left\{ \phi(o_N) + \sum_{t=0}^{N-1}L(o_t, a_t) \right\}\\
    \text{s.t.}\;\; &o_{t+1} = f(o_t, a_t), \;\; t = 0, 1, \dots, N-1\\
    &g_t(o_t, a_t)<0, \;\; t = 0, 1, \dots, N-1\\
    &g_N(o_N)<0
    \end{aligned}
    \label{eq:opt-cilqr}
\end{equation}
where $N$ is the planning horizon, $L(\cdot)$ and $\phi(\cdot)$ are the cost functions, $f(\cdot)$ is the transition model, and $g_t(\cdot)$'s are the safety and dynamics constraints. 

From a theoretical point of view, Chen \textit{et al} proved in Theorem 1 in \cite{CILQR} that for the problem in Equation~\ref{eq:opt-cilqr}, the output trajectories $\{ o_t^{(k)}, a_t^{(k)} \}$ at the $k-$th step will converge to a local optimum as $ k\rightarrow 0 $ when using the CILQR algorithm. Compared to other non-learning based methods, CILQR solves the optimal control problem with non-convex constraints and non-linear system dynamics much faster with a guarantee to converge. Learning based methods do not introduce constraints on dynamics and have no guarantee of a convergence either. 

It has been tested with on-road driving scenarios and is proved to be able to avoid obstacles successfully. However, the main drawback of CILQR is that it tends to be very aggressive if its reference speed is too fast. For example, when the ego agent is following slow traffic in an urban scenario, it always tries to pass the cars in the front whenever it finds a gap. This maneuver style may cause serious problems, because a sudden move is highly likely to result in collisions in such dense traffic with many occlusions. The dangerous decision is mainly because of the high reference speed that the controller tempts to track. Since the objective function penalizes its deviation from the reference trajectory, it is willing to sacrifice the safety to bypass the obstacles. 

Therefore, we seek to apply the high-level RL policy to choose the most suitable reference and the most appropriate setting for low-level controllers. We design a fixed finite set of candidate reference speed for the high-level RL to choose. The set includes 9 possible discrete speeds: $v_{\text{ref}}\in \{0, 2, 3, 4, 5, 6, 7, 8, 9\}(m/s)$. Each reference speed corresponds to a different CILQR controller that has a different behavior pattern. For example, the one with the reference speed $v_{\text{ref}} = 0m/s$ is the most conservative one, because it would yield to any obstacle in the environment, whereas the one with $v_{\text{ref}} = 9m/s$ is the most aggressive one, for it would tries its best to track the high reference speed and the safety would be compromised. The high-level RL aims to balance between the safety and the passing time given the current observations, so as to make the ego agent capable of handling various complex scenarios. 

\subsection{Baseline Methods}
\begin{figure*}[thpb]
     \centering
     \begin{subfigure}[b]{0.3\textwidth}
         \centering
         \includegraphics[width=\textwidth]{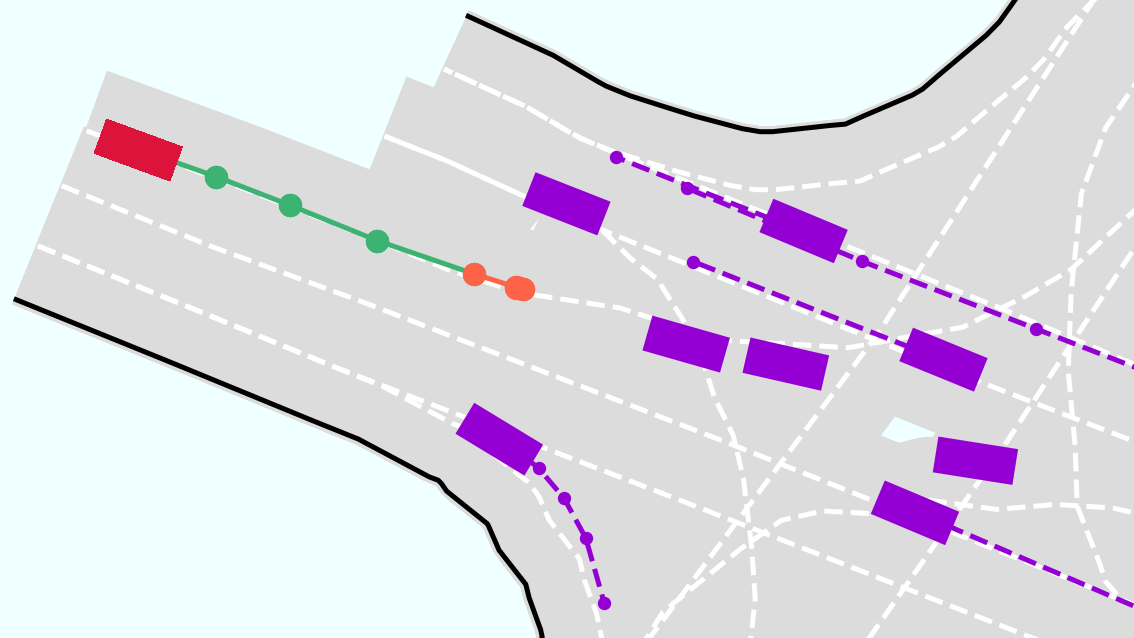}
         \caption{VA: 1-Entering}
         \label{fig:tc1}
     \end{subfigure}
     \hfill
     \begin{subfigure}[b]{0.3\textwidth}
         \centering
         \includegraphics[width=\textwidth]{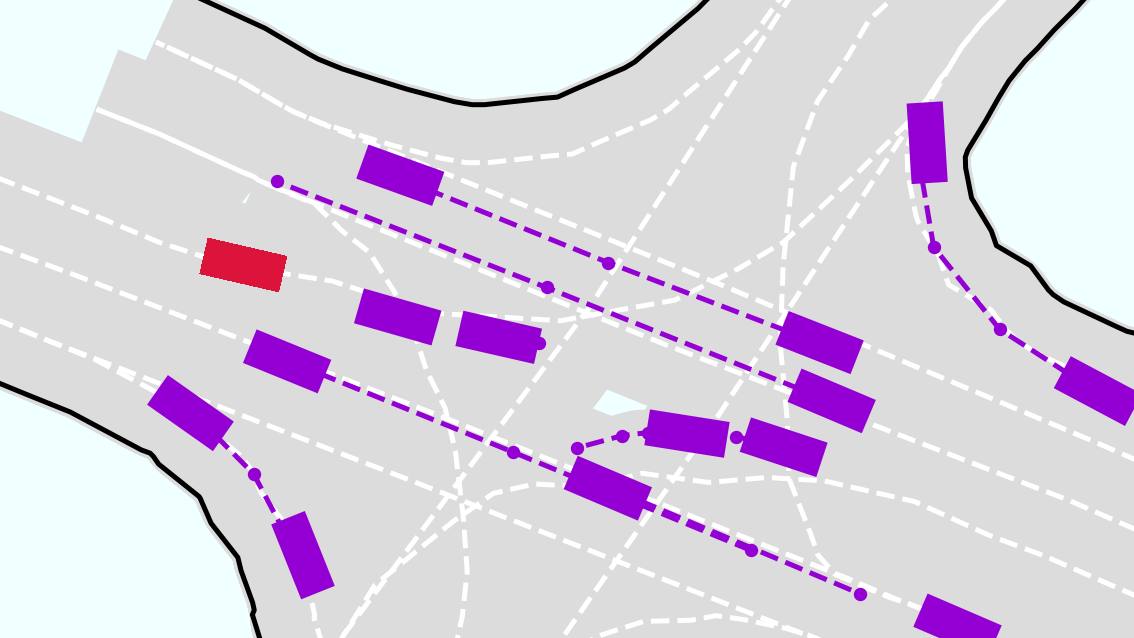}
         \caption{VA: 2-Waiting}
         \label{fig:tc2}
     \end{subfigure}
     \hfill
     \begin{subfigure}[b]{0.3\textwidth}
         \centering
         \includegraphics[width=\textwidth]{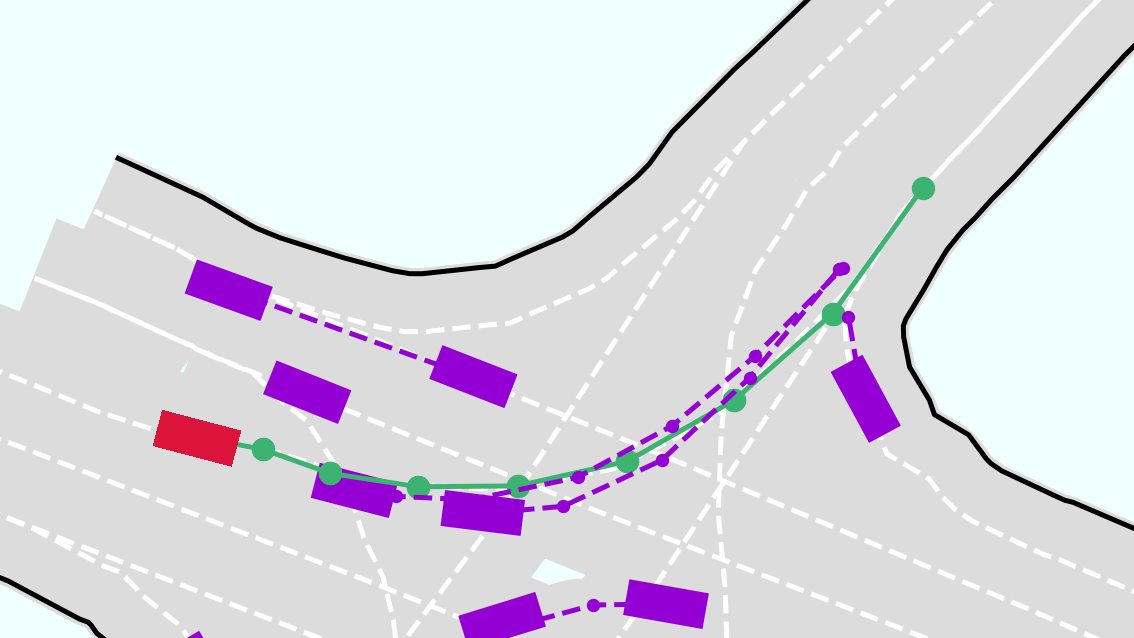}
         \caption{VA: 3-Exiting}
         \label{fig:tc3}
     \end{subfigure}
     \hfill
     \begin{subfigure}[b]{0.3\textwidth}
         \centering
         \includegraphics[width=\textwidth]{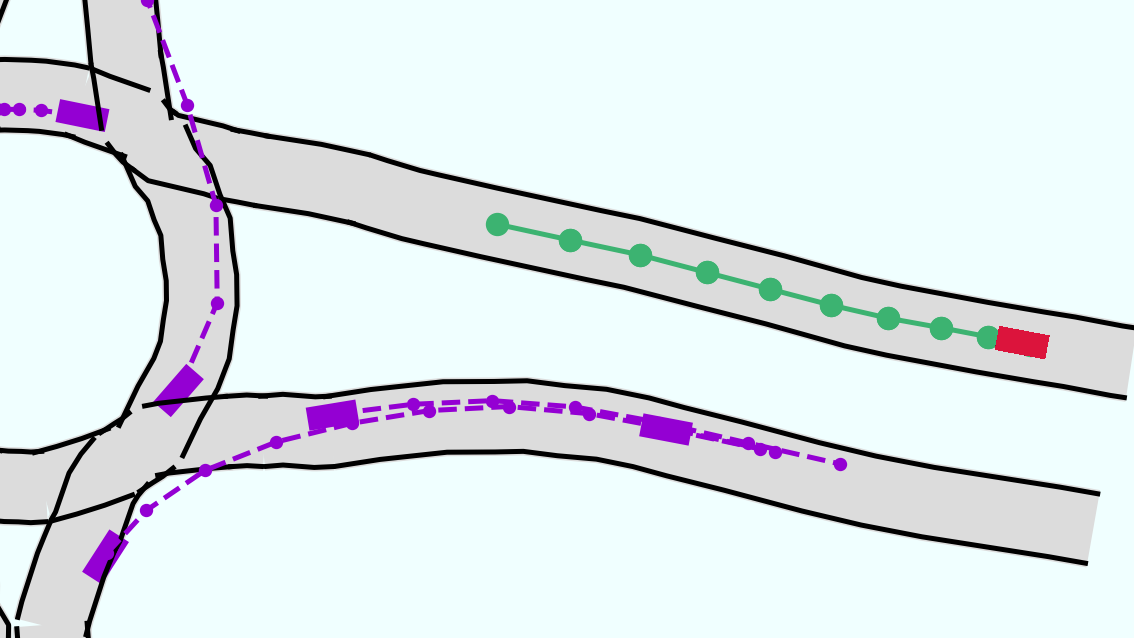}
         \caption{SR: 1-Entering}
         \label{fig:sr1}
     \end{subfigure}
     \hfill
     \begin{subfigure}[b]{0.3\textwidth}
         \centering
         \includegraphics[width=\textwidth]{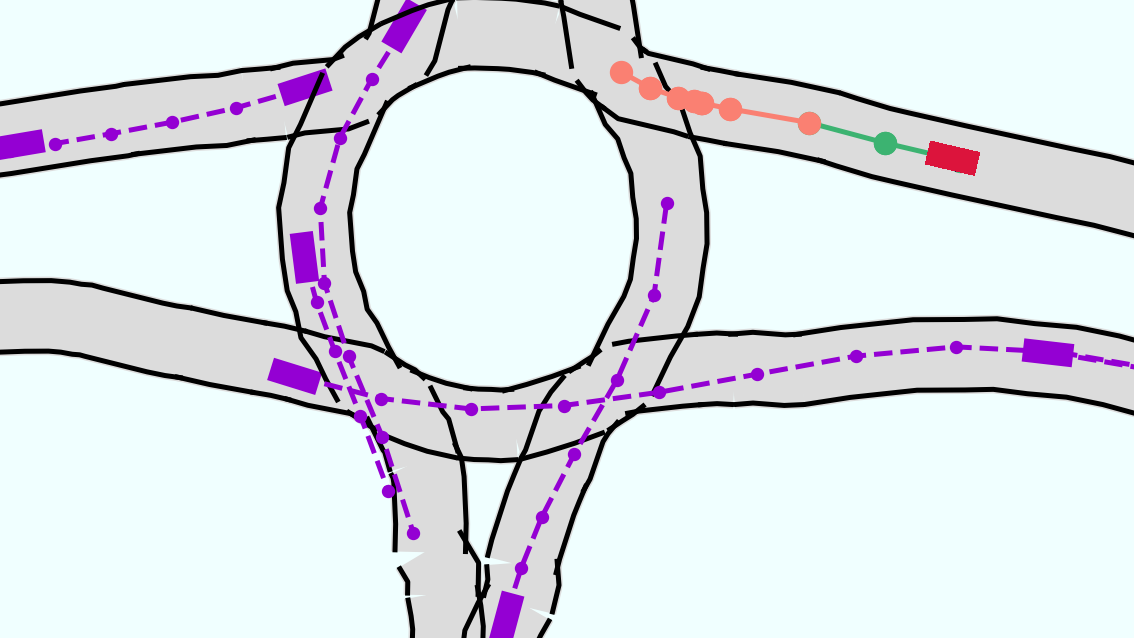}
         \caption{SR: 2-Waiting}
         \label{fig:sr2}
     \end{subfigure}
     \hfill
     \begin{subfigure}[b]{0.3\textwidth}
         \centering
         \includegraphics[width=\textwidth]{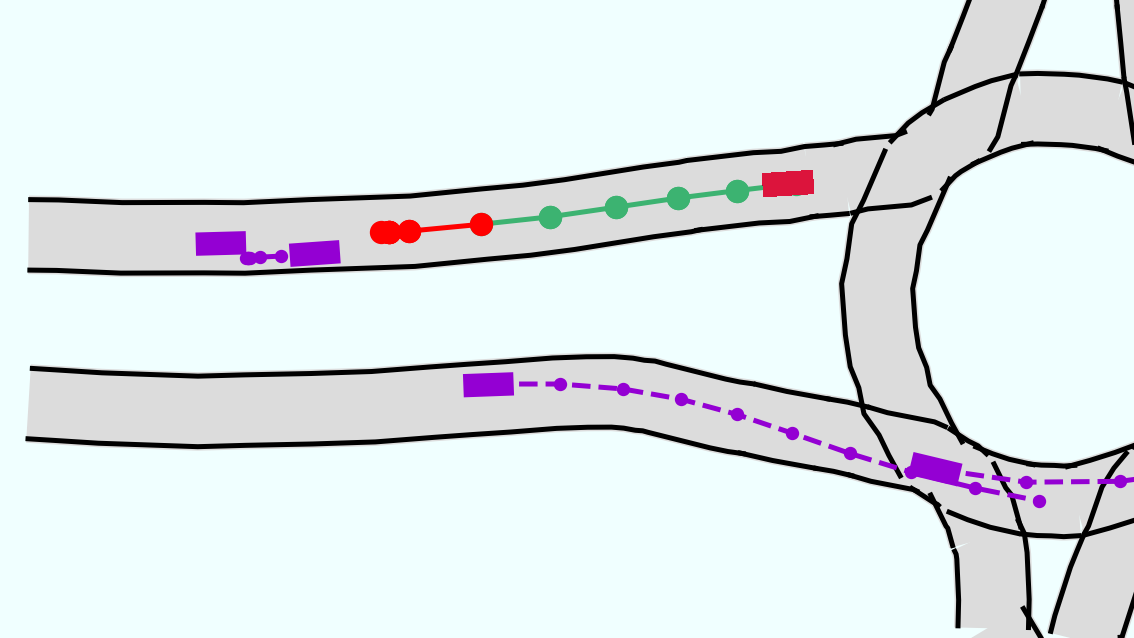}
         \caption{SR: 3-Following}
         \label{fig:sr3}
     \end{subfigure}
        \caption{The planned trajectories in both maps. The ego car is always red and the obstacles are always purple. The future trajectories for the next 10 high-level timestep $h$ are plotted using a line with markers on it. A green line means H-CtRL chooses a low-level CILQR planner with high reference speed, while a red line means H-CtRL chooses one with low reference speed. The darker the red, the lower the speed.}
        \label{fig:path-results}
\end{figure*}

We compared the following policies in the experiments:
\begin{itemize}
    \item \textbf{CILQR\#3}. The third low-level CILQR controller with a reference speed of $3m/s$. No high-level RL policy is used.
    \item \textbf{CILQR\#9}. The ninth low-level CILQR controller with a reference speed of $9m/s$. No high-level RL policy is used.
    \item \textbf{Random}. The hierarchical framework with the high-level RL is replaced by a random sampler, which chooses low-level controllers randomly.
    \item \textbf{H-CtRL}. Our proposed hierarchical behavior planning framework.
\end{itemize}
It would be tedious to compare and list results of all low-level CILQR controllers, so we only choose two representative low-level controllers here. CILQR\#3 plans trajectory that tracks a low reference speed, which will results in a rather conservative driving strategy, whereas CILQR\#9 is just the opposite resulting in an aggressive driving strategy. 

\subsection{Implementation Details}

We trained and tested the RL in our proposed hierarchical framework in two maps separately. 
At the beginning of each episode, we initialized the position of the ego vehicle at the edge of the road map, perturbed by a Gaussian noise. The initial timestep to load obstacles from the dataset was randomly sampled from $600$ to $900$ original timestep in VA and from $100$ to $400$ in SR. Each timestep in the simulator as well as CILQRs lasts $0.1$s, whereas each timestep of the high-level RL lasts $1.0$s with $p=10$. 

The goal of each episode was also chosen randomly, either to turn left, turn right, or to go straight. The observation that was fed into the high-level RL was the observation given by the simulator plus the goal of each episode. The RL policy was represented by a neural network with two fully connected hidden layers. According to the high-level decision, the same observation was then fed into the low-level controller to plan executable actions in the simulator for the ego agent.

\begin{center}
\begin{table}[]
    \centering
    \caption{Average episode returns, collision rate, and the completion rate within 50s time limit  based on 100 episodes in each map. }
    \begin{tabular}{c c c c}
    \Xhline{4\arrayrulewidth}\\[-1em]
    Method & Aver. Epi. Return & Collision Rate & Completion Rate \\
    \hline\\[-1em]
    \multicolumn{4}{l}{\textbf{Intersection (VA)}}\\[2pt]
    CILQR\#3 & 80.06 & 0.07 & 0.24 \\
    CILQR\#9 & 37.84 & 0.59 & 0.09 \\
    Random   & 51.23 & 0.36 & 0.17 \\
    H-CtRL   & 86.59 & 0.10 & 0.85 \\
    \hline\\[-1em]
    \multicolumn{4}{l}{\textbf{Roundabout (SR)}}\\[2pt]
    CILQR\#3 & 73.52 & 0.05 & 0.17 \\
    CILQR\#9 & 48.64 & 0.47 & 0.22 \\
    Random   & 55.13 & 0.32 & 0.28 \\
    H-CtRL   & 90.29 & 0.08 & 0.91 \\
    \Xhline{4\arrayrulewidth}
    \end{tabular}
    \label{tab:results}
\end{table}
\end{center}

\section{RESULTS}

\subsection{Statistics}
We ran each policy for 100 episodes in VA and SR separately, and compared the average episode return, the collision rate, and the completion rate within a 50s time limit. 

As shown in Table \ref{tab:results}, the proposed H-CtRL has the best performance in terms of the average episode reward. It is much higher than CILQR\#9 and Random, while CILQR\#3 is close to it. Considering the experiment setting where initial states of the environment are sampled randomly among a wide range, we can safely conclude that our proposed H-CtRL is able to handle various situations better than each individual low-level planners. It also implies that H-CtRL is better than a random high-level switching policy, so the high-level RL successfully learned useful skills to navigate through various complex urban traffics. 

When looking into the collision rate, CILQR\#3 performs the best, following by H-CtRL. CILQR\#9 is the most aggressive policy, as is implied by its high reference speed. When we set a time limit of 50 seconds for each episode, only the proposed H-CtRL has the ability to finish the task in both maps. CILQR\#3 is too conservative and drives at a low speed, making it almost impossible to finish the task in time. CILQR\#9 is just the opposite of CILQR\#3. It drives too fast to recognize danger and to avoid collisions in time. The random switching policy fails to reach a high completion rate because of a mixture of the reasons aforementioned. If we look into the collision rate and the completion rate together, we can conclude that H-CtRL makes a good balance between the safety and the operation time. 

\subsection{Visualization}
To show the details of what happened in both VA and SR, we visualized one representative episode in each map. 

Figure \ref{fig:tc1} - \ref{fig:tc3} are the visualization for an episode in VA, where the ego car was trying to make a left turn when the traffic light was green. According to traffic rules, it must yield to oncoming cars from across the road. As we can see in Figure \ref{fig:tc1}, the high-level RL first chose to drive at a high speed of approximately $6m/s$ (as shown by the green markers) into the entrance of the intersection. Then it decided to slow down in front of the intersection (as shown by the orange markers). When cars continued to come from the opposite direction, the high-level RL planned a perfect stop to stay put (as shown in Figure \ref{fig:tc2}). After all cars finished crossing the intersection, It decided to accelerate and to exit the intersection as fast as possible (as shown in Figure \ref{fig:tc3}).

One episode in SR was visualized in Figure \ref{fig:sr1} - \ref{fig:sr3}. In this episode, the ego agent was trying to go straight across the roundabout. When it was approaching the roundabout, the high-level RL chose to drive at a high speed as usual (implied by the green markers in Figure \ref{fig:sr1}). When it was about to enter the roundabout, the high-level RL decided to slow down (shown in Figure \ref{fig:sr2}) so that it could observe the surroundings and could avoid collisions if necessary. After it exited the roundabout (shown in Figure \ref{fig:sr3}), it first accelerated toward the goal position. But there were several obstacles driving slowly in the front, so it chose a low-level controller with a very low reference speed ($v_{\text{ref}}=2m/s$). Then the low-level CILQR controller helped the ego agent to slow down and avoid collisions. 

\begin{figure}[thbp]
     \centering
     \begin{subfigure}[b]{0.46\textwidth}
         \centering
         \includegraphics[width=\linewidth]{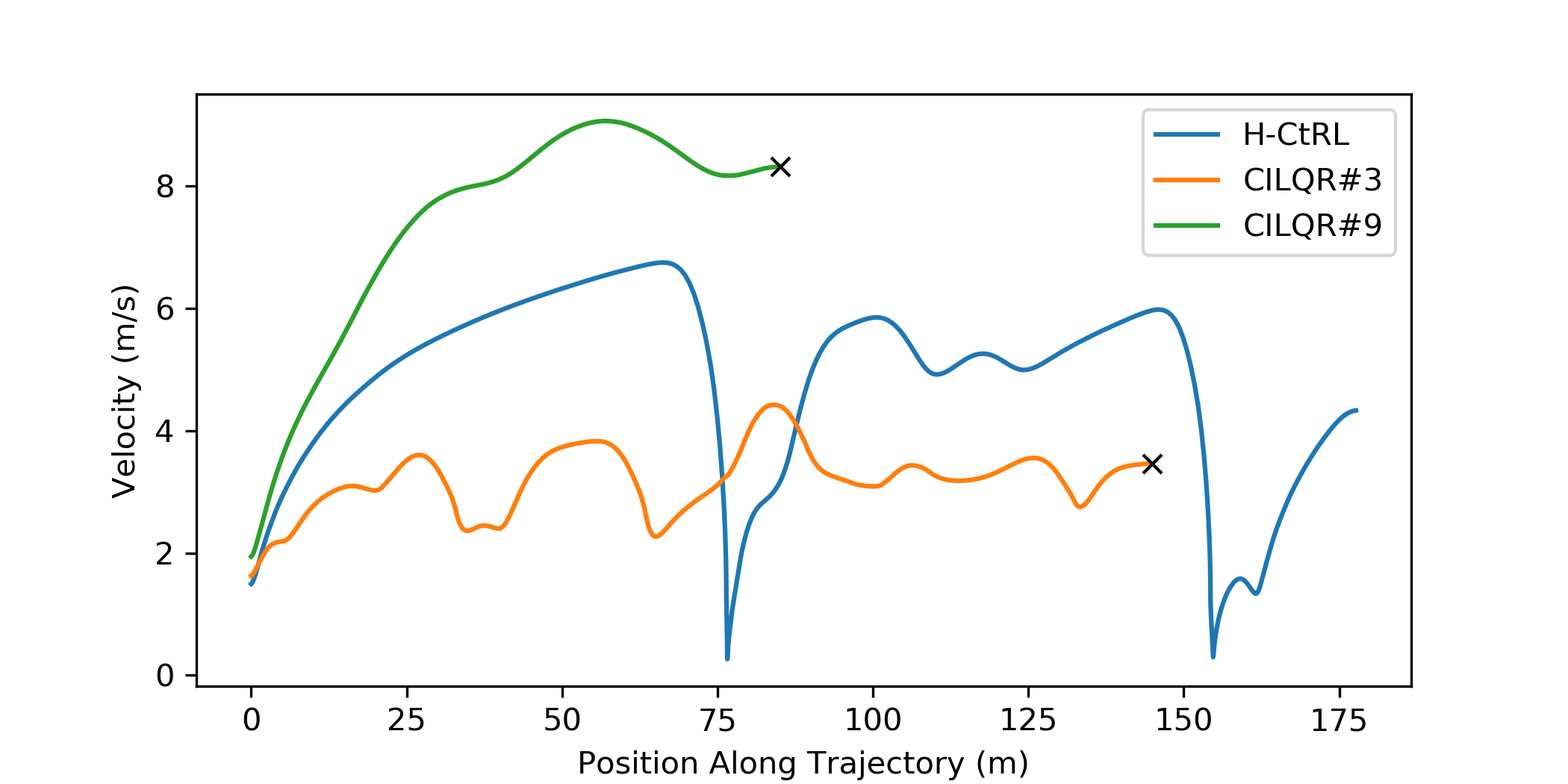}
         \caption{The velocity vs. the position along the trajectory of the ego agent.}
         \label{fig:velocity}
     \end{subfigure}
     \hfill
     \begin{subfigure}[b]{0.46\textwidth}
         \centering
         \includegraphics[width=\linewidth]{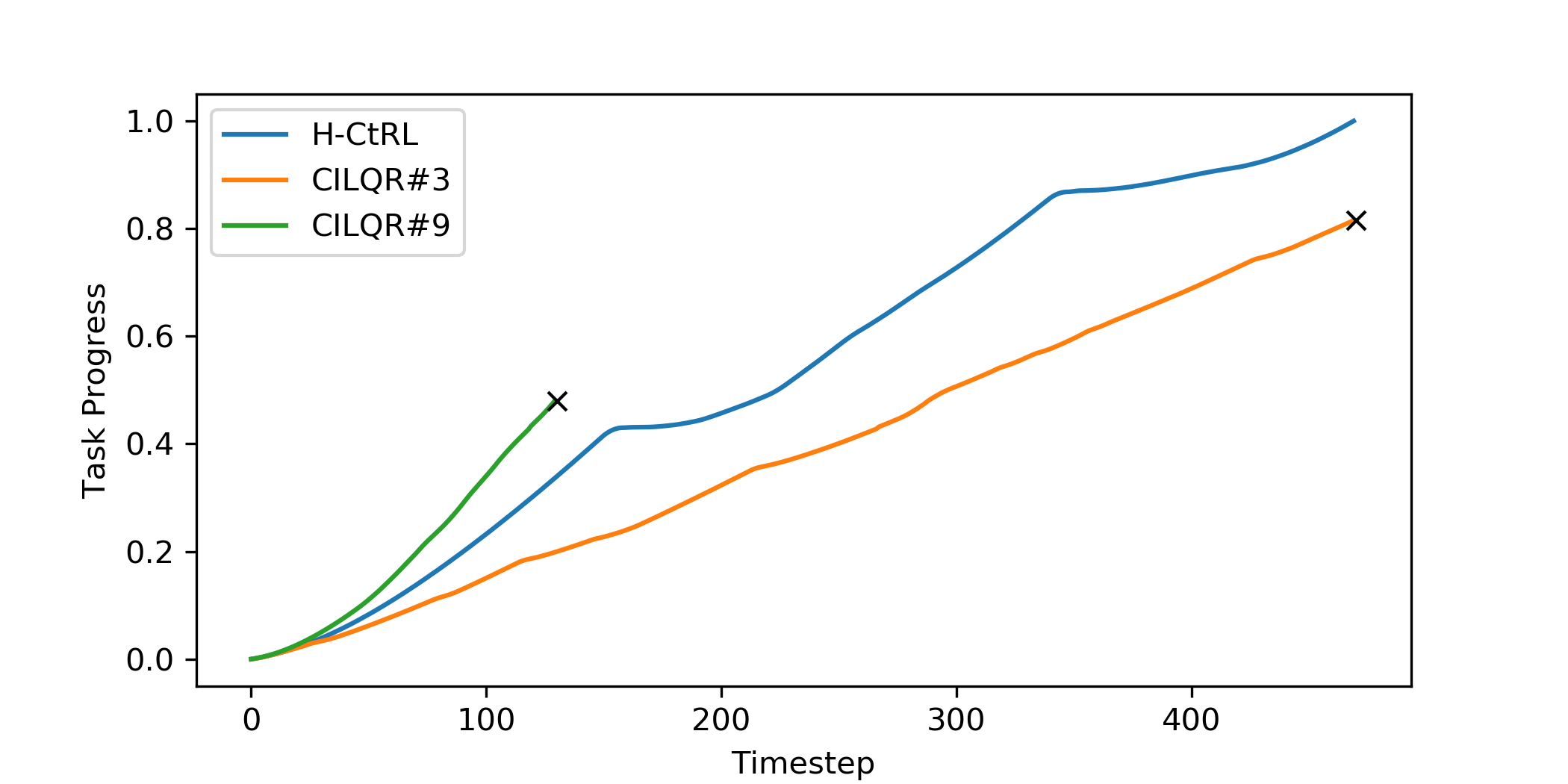}
         \caption{The task progress vs. the timestep of the ego agent.}
         \label{fig:progress}
     \end{subfigure}
        \caption{A visualization of the velocity and the task progress of the ego agent driving in SR. The proposed H-CtRL was compared to CILQR\#3 (with $v_\text{ref} = 3m/s$) and CILQR\#9 (with $v_\text{ref} = 9m/s$). The black cross means that the episode was terminated earlier without reaching the goal position. }
        \label{fig:lines}
\end{figure}

We visualized the velocity and the task progress of H-CtRL for the SR episode described above, and compared them with those of CILQR\#3 and CILQR\#9 in Figure \ref{fig:lines}. As we can see, neither CILQR\#3 nor CILQR\#9 managed to finish the task before the time was up. The ego agent with CILQR\#9 collided with obstacles in the episode, whereas the one with CILQR\#3 did not manage to finish the task before the time was up. Only the agent with H-CtRL successfully finished this episode. We can therefore conclude that our proposed H-CtRL has the ability to handle these complex urban traffic conditions safely and efficiently. 

\subsection{Failure Cases}
When training and testing our proposed method, we discovered that the ego agent braked really hard if it observed obstacles in the front or the front vehicle slowed down. We believed that this is caused by the low-level CILQR controllers, which only considered the safety constraint without any optimization on the comfort. Generally speaking, our proposed hierarchical framework has the ability to adopt many low-level planners. Therefore, the future work is to add more low-level planners into the framework, and to train them all together to get a more powerful and generalizable behavior planner.

\section{CONCLUSION}
In this paper, we proposed a general behavior planner for autonomous vehicles based on reinforcement learning and safe motion planners. By combining the power of low-level safe controllers with a high-level reinforcement learning coordinator, various complex urban traffic conditions can be handled via this general framework. 
We built a simulator that can reproduce scenarios according to real-world traffic dataset. The proposed algorithm was trained and tested based on such real traffic data. Compared to other baseline methods, the proposed framework achieved both high completion rate and low collision rate, verifying its ability to handle various traffic scenarios with satisfying performance on both safety and efficiency.

\addtolength{\textheight}{-9.5cm}   


\bibliographystyle{ieeetr}
\bibliography{reference.bib}

\begin{thebibliography}{10}

\bibitem{mcnaughton2011motion}
M.~McNaughton, C.~Urmson, J.~M. Dolan, and J.-W. Lee, ``Motion planning for
  autonomous driving with a conformal spatiotemporal lattice,'' in {\em 2011
  IEEE International Conference on Robotics and Automation}, pp.~4889--4895,
  IEEE, 2011.

\bibitem{paden2016survey}
B.~Paden, M.~{\v{C}}{\'a}p, S.~Z. Yong, D.~Yershov, and E.~Frazzoli, ``A survey
  of motion planning and control techniques for self-driving urban vehicles,''
  {\em IEEE Transactions on intelligent vehicles}, vol.~1, no.~1, pp.~33--55,
  2016.

\bibitem{defensive}
W.~Zhan, C.~Liu, C.-Y. Chan, and M.~Tomizuka, ``A non-conservatively defensive
  strategy for urban autonomous driving,'' pp.~459--464, 11 2016.

\bibitem{bayesianPersuasive}
C.~{Peng} and M.~{Tomizuka}, ``Bayesian persuasive driving,'' in {\em 2019
  American Control Conference (ACC)}, pp.~723--729, 2019.

\bibitem{leverageHuman}
D.~Sadigh, S.~Sastry, S.~A. Seshia, and A.~D. Dragan, ``Planning for autonomous
  cars that leverage effects on human actions.,'' in {\em Robotics: Science and
  Systems}, vol.~2, Ann Arbor, MI, USA, 2016.

\bibitem{CILQR}
J.~{Chen}, W.~{Zhan}, and M.~{Tomizuka}, ``Constrained iterative lqr for
  on-road autonomous driving motion planning,'' in {\em 2017 IEEE 20th
  International Conference on Intelligent Transportation Systems (ITSC)},
  pp.~1--7, 2017.

\bibitem{Alvinn}
D.~A. Pomerleau, ``Alvinn: An autonomous land vehicle in a neural network,'' in
  {\em Advances in neural information processing systems}, pp.~305--313, 1989.

\bibitem{end2endforAV}
M.~Bojarski, D.~D. Testa, D.~Dworakowski, B.~Firner, B.~Flepp, P.~Goyal, L.~D.
  Jackel, M.~Monfort, U.~Muller, J.~Zhang, X.~Zhang, J.~Zhao, and K.~Zieba,
  ``End to end learning for self-driving cars,'' 2016.

\bibitem{endtoendPerception}
H.~Xu, Y.~Gao, F.~Yu, and T.~Darrell, ``End-to-end learning of driving models
  from large-scale video datasets,'' 2017.

\bibitem{sun2018fast}
L.~Sun, C.~Peng, W.~Zhan, and M.~Tomizuka, ``A fast integrated planning and
  control framework for autonomous driving via imitation learning,'' in {\em
  Dynamic Systems and Control Conference}, vol.~51913, p.~V003T37A012, American
  Society of Mechanical Engineers, 2018.

\bibitem{DILinurban}
J.~Chen, B.~Yuan, and M.~Tomizuka, ``Deep imitation learning for autonomous
  driving in generic urban scenarios with enhanced safety,'' 2019.

\bibitem{DRLframework}
A.~Sallab, M.~Abdou, E.~Perot, and S.~Yogamani, ``Deep reinforcement learning
  framework for autonomous driving,'' {\em Electronic Imaging}, vol.~2017,
  p.~70–76, Jan 2017.

\bibitem{virtualtoreal}
X.~Pan, Y.~You, Z.~Wang, and C.~Lu, ``Virtual to real reinforcement learning
  for autonomous driving,'' 2017.

\bibitem{scalable}
M.~{Bouton}, A.~{Nakhaei}, K.~{Fujimura}, and M.~J. {Kochenderfer}, ``Scalable
  decision making with sensor occlusions for autonomous driving,'' in {\em 2018
  IEEE International Conference on Robotics and Automation (ICRA)},
  pp.~2076--2081, 2018.

\bibitem{interpretable}
J.~Chen, S.~E. Li, and M.~Tomizuka, ``Interpretable end-to-end urban autonomous
  driving with latent deep reinforcement learning,'' 2020.

\bibitem{evolvegraph}
J.~Li, F.~Yang, M.~Tomizuka, and C.~Choi, ``Evolvegraph: Multi-agent trajectory
  prediction with dynamic relational reasoning,'' in {\em 2020 Advances in
  Neural Information Processing Systems (NeurIPS)}, 2020.

\bibitem{conditional}
J.~Li, H.~Ma, and M.~Tomizuka, ``Conditional generative neural system for
  probabilistic trajectory prediction,'' in {\em 2019 IEEE/RSJ International
  Conference on Intelligent Robots and Systems (IROS)}, pp.~6150--6156, IEEE,
  2019.

\bibitem{HRL}
Z.~Qiao, Z.~Tyree, P.~Mudalige, J.~Schneider, and J.~M. Dolan, ``Hierarchical
  reinforcement learning method for autonomous vehicle behavior planning,''
  2019.

\bibitem{towardshrl}
M.~S. Nosrati, E.~A. Abolfathi, M.~Elmahgiubi, P.~Yadmellat, J.~Luo, Y.~Zhang,
  H.~Yao, H.~Zhang, and A.~Jamil, ``Towards practical hierarchical
  reinforcement learning for multi-lane autonomous driving,'' 2018.

\bibitem{learningHRL}
J.~Wang, Y.~Wang, D.~Zhang, Y.~Yang, and R.~Xiong, ``Learning hierarchical
  behavior and motion planning for autonomous driving,'' 2020.

\bibitem{hoel2019combining}
C.-J. Hoel, K.~Driggs-Campbell, K.~Wolff, L.~Laine, and M.~J. Kochenderfer,
  ``Combining planning and deep reinforcement learning in tactical decision
  making for autonomous driving,'' {\em IEEE Transactions on Intelligent
  Vehicles}, vol.~5, no.~2, pp.~294--305, 2019.

\bibitem{thananjeyan2020safety}
B.~Thananjeyan, A.~Balakrishna, U.~Rosolia, F.~Li, R.~McAllister, J.~E.
  Gonzalez, S.~Levine, F.~Borrelli, and K.~Goldberg, ``Safety augmented value
  estimation from demonstrations (saved): Safe deep model-based rl for sparse
  cost robotic tasks,'' {\em IEEE Robotics and Automation Letters}, vol.~5,
  no.~2, pp.~3612--3619, 2020.

\bibitem{rl-il}
Z.~Cao, E.~Bıyık, W.~Z. Wang, A.~Raventos, A.~Gaidon, G.~Rosman, and
  D.~Sadigh, ``Reinforcement learning based control of imitative policies for
  near-accident driving,'' 2020.

\bibitem{socialAttention}
E.~Leurent and J.~Mercat, ``Social attention for autonomous decision-making in
  dense traffic,'' 2019.

\bibitem{bicyclemodel}
J.~{Kong}, M.~{Pfeiffer}, G.~{Schildbach}, and F.~{Borrelli}, ``Kinematic and
  dynamic vehicle models for autonomous driving control design,'' in {\em 2015
  IEEE Intelligent Vehicles Symposium (IV)}, pp.~1094--1099, 2015.

\bibitem{ddqn}
H.~van Hasselt, A.~Guez, and D.~Silver, ``Deep reinforcement learning with
  double q-learning,'' 2015.

\bibitem{interactiondataset}
W.~Zhan, L.~Sun, D.~Wang, H.~Shi, A.~Clausse, M.~Naumann, J.~K\"ummerle,
  H.~K\"onigshof, C.~Stiller, A.~de~La~Fortelle, and M.~Tomizuka,
  ``{INTERACTION} {Dataset}: {An} {INTERnational}, {Adversarial} and
  {Cooperative} {moTION} {Dataset} in {Interactive} {Driving} {Scenarios} with
  {Semantic} {Maps},'' {\em arXiv:1910.03088 [cs, eess]}, 2019.

\end{thebibliography}


\end{document}